\documentclass[final]{cvpr}

\usepackage{times}
\usepackage{epsfig}
\usepackage{graphicx}
\usepackage{amsmath}
\usepackage{amssymb}

\usepackage[pagebackref=true,breaklinks=true,colorlinks,bookmarks=false]{hyperref}

\def\cvprPaperID{2349} 
\def\confYear{CVPR 2021}

\begin{document}


\title{Distractor-Aware Fast Tracking via Dynamic Convolutions and MOT Philosophy}

\newcommand*{\affaddr}[1]{#1} 
\newcommand*{\affmark}[1][*]{\textsuperscript{#1}}
\newcommand*{\email}[1]{\texttt{#1}}

\author{
Zikai Zhang\affmark[1,2], Bineng Zhong\affmark[1]\thanks{Corresponding author.}, Shengping Zhang\affmark[3,4], Zhenjun Tang\affmark[1], Xin Liu\affmark[5], Zhaoxiang Zhang\affmark[6]\\
\affaddr{\affmark[1]Guangxi Key Lab of Multi-Source Information Mining \& Security, Guangxi Normal University},\\
\affaddr{\affmark[2]Department of Computer Science and Technology, Huaqiao University},\\
\affaddr{\affmark[3]Harbin Institute of Technology},
\affaddr{\affmark[4]}Peng Cheng Laboratory,
\affaddr{\affmark[5]Beijing Seetatech Technology},\\
\affaddr{\affmark[6]Institute of Automation, CAS \& University of Chinese Academy of Sciences \\ \& Centre for Artificial Intelligence and Robotics, HKISI\_CAS}\\
\email{zikaizhang@hqu.edu.cn},
\email{bnzhong@gxnu.edu.cn},
\email{s.zhang@hit.edu.cn}\\
\email{tangzj230@163.com},
\email{xin.liu@seetatech.com},
\email{zhaoxiang.zhang@ia.ac.cn}\\
}

\maketitle

\pagestyle{empty}
\thispagestyle{empty}

\begin{abstract}

A practical long-term tracker typically contains three key properties, \ie an efficient model design, an effective global re-detection strategy and a robust distractor awareness mechanism. However, most state-of-the-art long-term trackers (\eg, Pseudo and re-detecting based ones) do not take all three key properties into account and therefore may either be time-consuming or drift to distractors. To address the issues, we propose a two-task tracking framework (named {\bf DMTrack}), which utilizes two core components (\ie, one-shot detection and re-identification (re-id) association) to achieve distractor-aware fast tracking via {\bf D}ynamic convolutions (d-convs) and {\bf M}ultiple object tracking (MOT) philosophy. To achieve precise and fast global detection, we construct a lightweight one-shot detector using a novel dynamic convolutions generation method, which provides a unified and more flexible way for fusing target information into the search field. To distinguish the target from distractors, we resort to the philosophy of MOT to reason distractors explicitly by maintaining all potential similarities' tracklets. Benefited from the strength of high recall detection and explicit object association, our tracker achieves state-of-the-art performance on the LaSOT, OxUvA, TLP, VOT2018LT and VOT2019LT benchmarks and runs in real-time (3x faster than comparisons)\footnote{The code will be available at \url{https://github.com/hqucv/dmtrack}}.

\end{abstract}

\begin{figure}
\begin{center}
\setlength{\fboxrule}{0pt}
\setlength{\fboxsep}{0cm}
\fbox{\includegraphics[width=1\linewidth]{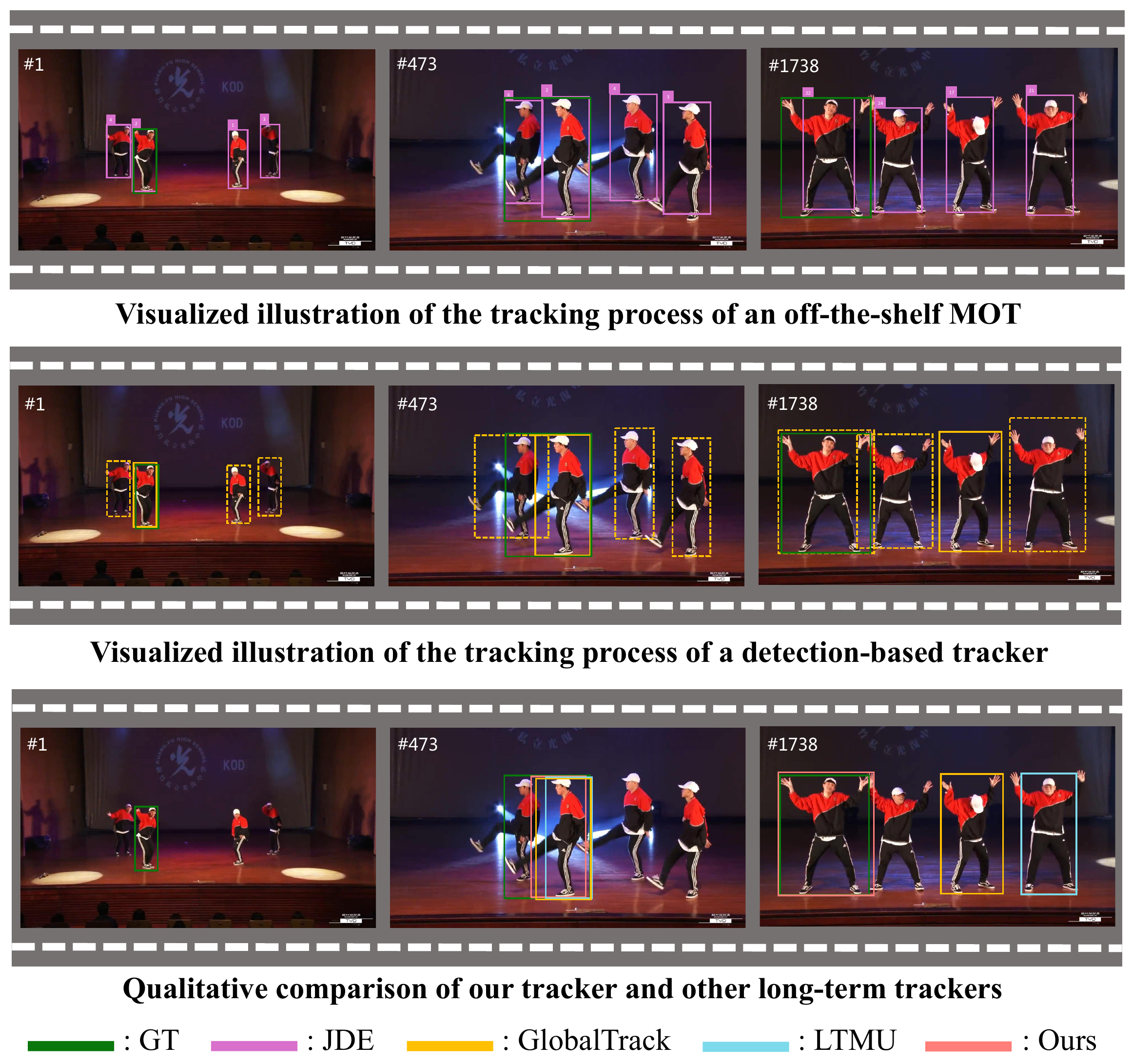}}
\end{center}
  \caption{Visualization of long-term tracking results on \textit{person-5} from LaSOT~\cite{lasot}. ``GT" means ground truth. ``GlobalTrack~\cite{globaltrack}" and ``LTMU~\cite{ltmu}" are two strong long-term trackers. ``JDE~\cite{jde}" is a multi-object tracker. In the first line, we show long-term tracking results from the off-the-shelf MOT model\cite{jde}. The object ids are in the upper left corner of the bounding boxes. In the second line, we show top-4 classification results from a detection-based tracker\cite{globaltrack}. The solid lines show the top-1 predictions. In the third line, we compare our DMTrack with state-of-the-art comparison, and present that distractor awareness is vital to visual object trackers. Better viewed in color with zoom-in.}
\label{fig:head}
\end{figure}

\section{Introduction}
Visual object tracking has drawn great attention to large-scale long-term tracking because of its great potential in real-world applications. The main difference between long-term and short-term trackers is that the former has to deal with the cases in which the target disappears and reappears frequently. Generally, long-term tracking sequences~\cite{lasot, tlp, oxuva} last for hundreds and thousands frames, which usually contain challenges such as appearance change, long-duration disappearance and intra-class distractors. Therefore, long-term trackers should have the abilities of re-detecting objects effectively and distinguish the target from similar distractors (As shown in Figure~\ref{fig:head}).

Recently, a large number of long-term trackers have been proposed~\cite{ltmu, splt, mbmd, siamdw}. Luke{\'z}i{\v{c}} \etal~\cite{vot2020-1} group long-term trackers into two categorizations: Pseudo long-term tracker (LT$_0$) and re-detecting long-term tracker (LT$_1$). LT$_0$ applies some short-term trackers~\cite{srdcf, bn-tracking, ocean} to the long-term tracking task straightforwardly by simply using the classification score to distinguish the target from its background. However, these trackers are prone to drift to distractors due to appearance confusion. LT$_1$ (e.g., SiamDW\_LT~\cite{siamdw}, MBMD~\cite{mbmd}, SPLT~\cite{splt}, LTMU~\cite{ltmu}) uses a re-detection strategy to recover from tracking failure. However, these trackers require a sophisticated design for interaction between local trackers and global detectors. Recently, Huang \etal~\cite{globaltrack} propose a global instance search (GIS) based tracker using a two-stage detector without motion constraint. Under the one-shot detection scheme, long-term tracking is simplified because the switch strategy that used for balancing the local and global modules is no longer needed. However, the heavy computing burdens and unstable performance caused by global detection make GIS-based methods improper for real-world applications.

To address the above issues, we propose a two-task tracking framework, which consists of a lightweight detection model and an explicit object association method. For the first task, we reach back to the correlation methods between the template and the search field in tracking and unify these methods into a dynamic convolutions (d-convs) generation paradigm~\cite{condconv}. Given powerful dynamic convolutions, we can embed target information into a one-stage anchor-free detection model with multiple kernel designs and integration layers while requiring less computation. For the second task, we resort to the philosophy of multiple object tracking (MOT). Specifically, we introduce a novel re-id embedding into the above detection model by jointly learning the two tasks. Benefiting from the discriminative re-id features, our tracker achieves favorable performance with a compact association strategy. The two tasks are implemented in the MOT framework which reasons distractors explicitly by maintaining all potential similarities tracklets. Experiments show that our tracker achieves state-of-the-art performance on the five long-term benchmarks and runs 3x faster than comparisons. 

Our main contributions can be summarized as follows,

\begin{itemize}
\item We propose a two-task long-term tracking framework, which contains a lightweight detector and an explicit object association. By implementing the two tasks in the MOT framework, our tracker obtains a fast inference speed and can distinguish the target from the distractors.
\item To build a high-efficiency detector, we present a novel dynamic convolutions generation method. To avoiding drifting to distractors, we learn a discriminative re-id embedding to achieve effective tracklet association.
\item Our approach achieves state-of-the-art results on five long-term tracking benchmarks and runs in real-time, which shows that the proposed method can be a more practical baseline for GIS-based trackers.
\end{itemize}

\section{Related Work}

\subsection{Deep Long-Term Visual Tracking}
Deep learning-based models for long-term tracking have shown their great capability~\cite{ptav, oxuva, splt, mbmd, dasiam}. Recent top-ranked long-term trackers follow the local tracker and global re-detector schemes. MBMD~\cite{mbmd} combines regression and verification modules to the tracking framework, and use a sliding window strategy in image level for re-detecting. SPLT~\cite{splt} uses a skimming module to speed up the re-detect processing by skipping the certain regions. However, there is a tough problem in the local-global paradigm:  \textit{when to switch between the local tracker and the global re-detector?} However, some methods followed the pure re-detection paradigm, Huang \etal~\cite{globaltrack} proposed to track in global search scheme by introducing a two-stage anchor-based detection framework in tracking task. Voigtlaender \etal~\cite{siamr-cnn} further use a cascade detection head for precise results. However, the computing burdens are heavy in these methods. In this paper, we develop an efficient detection model under one-stage anchor-free paradigm, which gets a favorable balance between speed and accuracy.

\subsection{Distractor Problem in Visual Tracking}
The ability of dealing with similar objects is important for long-term tracker. However, distractor problem is ill-posed due to the dynamic interaction of target and distractors. Zhu \etal~\cite{dasiam} propose an incremental learning method for online distractor suppression. Voigtlaender \etal~\cite{siamr-cnn} implement a hard example mining to guide model learning and use a dynamic programming algorithm to consistently suppress potential distractors.
Nevertheless, these methods are burdensome 
for accurate and efficient tracking. In this work, we address distractor problem by explicitly tracking all the potential objects in the MOT framework. Inspired by the joint learning methods of MOT~\cite{jde, fairmot, bn-reid}, we design a compact two-task tracking framework with one-shot detection and re-id association that runs in real-time.

\subsection{Correlation Methods for Visual Tracking}
Correlation operations that used for fusing the template information and the search field are seldom discussed. Bertinetto \etal~\cite{siamfc} use a simple cross-correlation operation to generate a similarity scoring map. Li \etal~\cite{siamrpn++} proposed a depthwise cross-correlation to reduce the computational cost. Huang \etal~\cite{globaltrack} use the Hadamard production to encode correlation information. However, these methods are restricted to fixed model structures. Contrast to fixed convolutional layer designs, d-convs are dynamically generated by using some conditional information. Dynamic filter network~\cite{dynamicfilter} and CondConv~\cite{condconv} explore the power of d-convs for increasing the capacity of the classification model. CondInst~\cite{condinst} and SOLOv2~\cite{solov2} use conditional convolution to embed position information into the mask branch for boosting the segmentation performance. Here, we utilize a dynamic convolutions generation method for feature correlation.

\section{Method}
\begin{figure*}
\begin{center}
\setlength{\fboxrule}{0pt}
\setlength{\fboxsep}{0cm}
\fbox{\includegraphics[width=.95\linewidth]{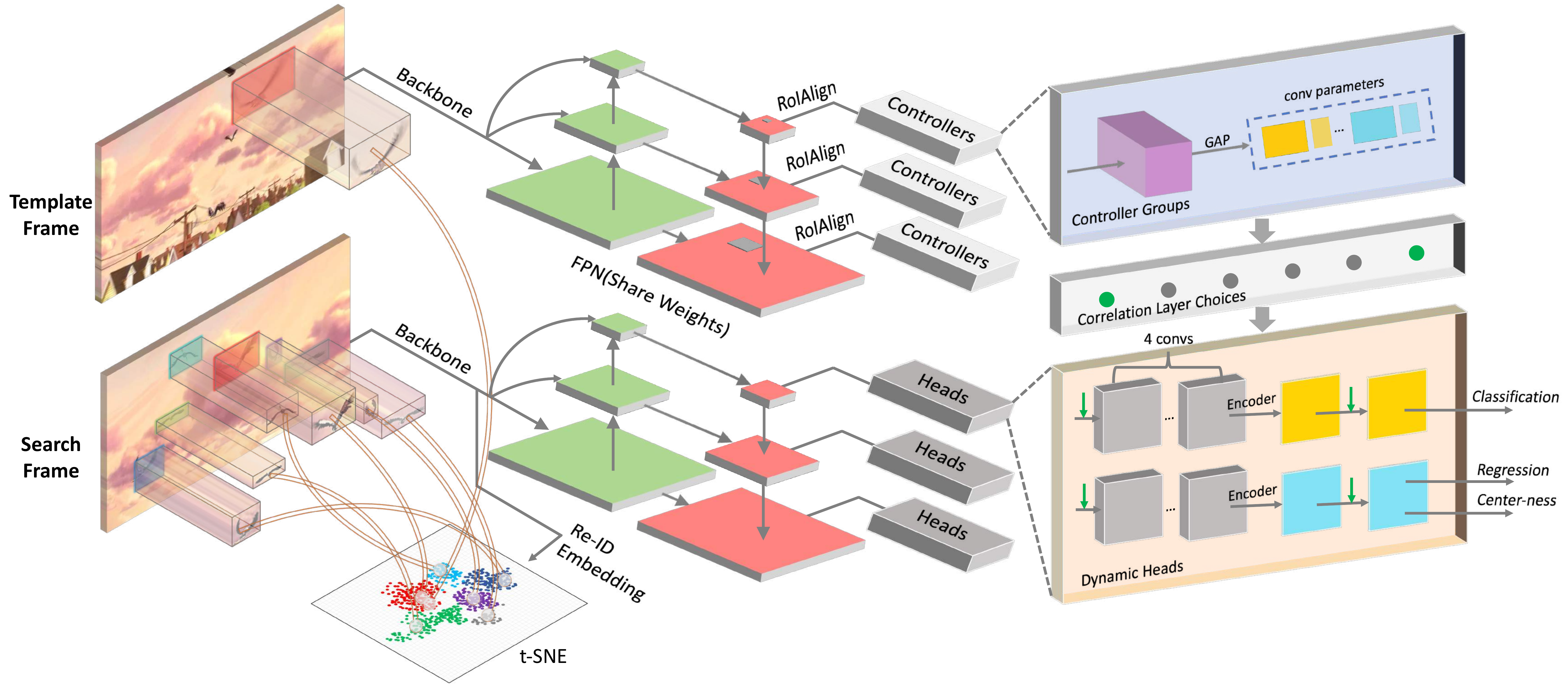}}
\end{center}
  \caption{The overall architecture of our model. The framework consists of three main components: a template branch for dynamic convolutions generation, a search branch for efficient one-shot detection and a re-id embedding for object association.}
\label{fig:pipeline}
\end{figure*}

In this section, we unveil the power of GIS-based trackers~\cite{gapdt, globaltrack, siamr-cnn} by designing a lightweight detector and a re-id embedding with capable association strategy. As shown in Figure~\ref{fig:pipeline}, our DMTrack consists of a group of dynamic convolution controllers, an efficient one-shot detection branch and a re-id embedding.

\subsection{Motivation}
\label{section:motivation}
\textit{What is the strength of global search scheme for long-term tracking? How can we achieve a more stable global tracker?} We dive into GIS-based methods, and answer these two questions by designing experiments to analysis the capabilities of modern trackers. For the first question, we show GIS-based methods have a high-quality proposal generating ability. And for the second question, we demonstrate the importance of the association ability by using an off-the-shelf MOT model to evaluate on a single object tracking benchmark. These two experiments are meant to show important factors that we must take into account when building a practical GIS-based tracker.

\noindent
{\bf Proposal Generator.} High-quality proposals are important for a tracking system. Though local search-based trackers achieve favorable performance in short-term scenes, most of them degenerate in large-scale long-term benchmarks~\cite{lasot, vot2019result, tlp, oxuva}. Therefore, a global re-detector becomes a core component for long-term tracking. Actually, we can treat a global re-detector as a proposal generator. And a high-recall generator with only a few candidates will be beneficial to later tracking stages. 

\begin{figure}
\begin{center}
\setlength{\fboxrule}{0pt}
\setlength{\fboxsep}{0cm}
\fbox{\includegraphics[width=1\linewidth]{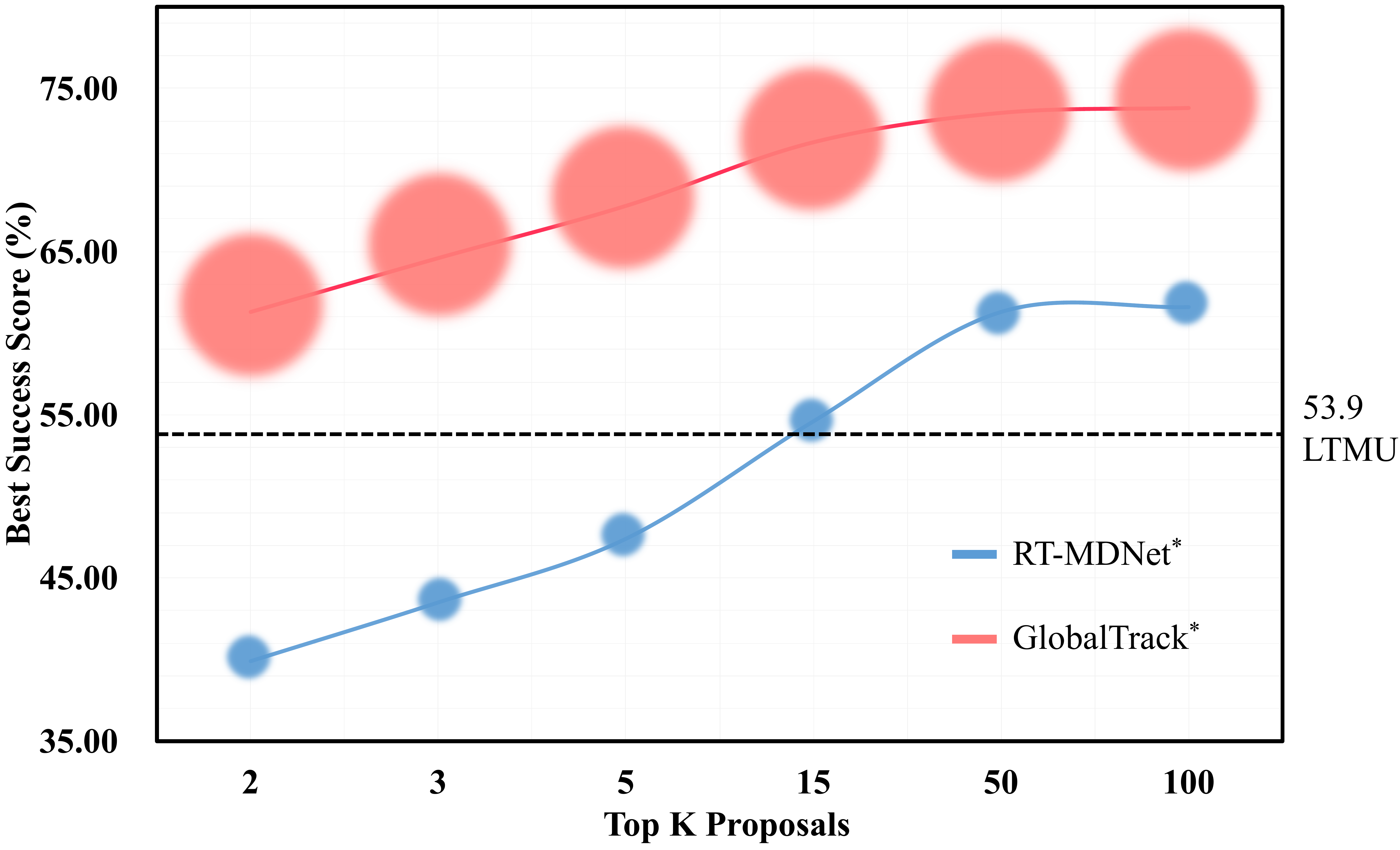}}
\end{center}
  \caption{Visualization of best success scores on LaSOT~\cite{lasot} and speeds from the local and global generator. The circle diameter means the relative time cost in inference stage.}
\label{fig:motivation}
\end{figure}

We experimentally evaluate the upper bounds of local and global trackers. Following popular protocol in OTB-2015~\cite{otb2015}, we perform the One-Pass Evaluation (OPE) and measure the best success score of top-K candidates of a local generator and a global generator. Here, we evaluate RT-MDNet~\cite{rtmdnet} as the local generator (marked as ``RT-MDNet$^*$") and GlobalTrack~\cite{globaltrack} as the global generator (marked as ``GlobalTrack$^*$"). For RT-MDNet, the tracker makes a gaussian sampling based on the position of the last frame prediction, we simply choose the top-k candidates by classification scores, then we determine the bounding box that has the highest intersection over Union score (IoU) as our output. For GlobalTrack, we just follow the identical strategy. Specifically, we implement the same classifier in these generators to control the variable. In the experiment, we set the candidate number K with 2, 3, 5, 15, 50, 100 and test on the LaSOT~\cite{lasot}. As shown in Figure ~\ref{fig:motivation}, GlobalTrack$^*$ exceed RT-MDNet$^*$ by a large margin. Even with only two candidates, the global-based generator achieves a success score of 61.3\%, which is comparable with the local-based generator with the top-50 score, and outperforms state-of-the-art long-term tracker LTMU~\cite{ltmu}. However, we can see that GIS-based method is time-consuming due to the heavy model design for global search (as shown in Figure~\ref{fig:motivation}). Therefore, one of the important factors that contribute to tracker's performance is a good balance between the precision and the efficiency.

\noindent
{\bf Off-the-Shelf MOT for Long-Term Tracking.} The above experiments have shown the capability of GIS-based trackers. In spite of the high proposal recall, the discrimination of these trackers is unsatisfactory. Intra-class objects reasoning is crucial for global search scheme. Here, we design an enlightening experiment to evaluate the long-term tracking performance of an off-the-shelf MOT model. Specifically, we choose sequences in long-term tracking benchmarks~\cite{lasot} that contains categories of \textit{person} (match with MOT pre-trained model). During inference stage, we maintain the object id that has the max IoU between the detections and the annotation in the first frame. In the subsequent frames, we just choose the prediction with the same id as our tracking result. In Figure~\ref{fig:head}, we present the tracking results of JDE~\cite{jde}. As we can see, MOT method keeps a robust and accurate tracking for the first hundred frames, which is surprising because the model is never trained on single object tracking dataset. Therefore, besides high-recall proposal generator, object association is also an important factor that carries a big weight in final tracking performance. 

Motivated by the above trials, we design a GIS-based tracker that considers computation cost and distractor awareness. In the next part, we make an overview on our DMTrack and demonstrate its core components. 

\subsection{Overall Architecture}
Given the template image $\mathit{I}^{\mathit{t}}\in\mathbb{R}^{H\times W\times 3}$ and search image $\mathit{I}^{\mathit{s}}\in\mathbb{R}^{H\times W\times 3}$, our tracking algorithm search the target in successive frames with only first frame annotation. In this work, we develop a GIS-based tracker that uses an anchor-free detection model as the base detection model. In order to build a class-agnostic detector, we introduce the dynamic convolutions controller to generate convolution parameters conditioned on the target information. Further, we jointly learn the detection task with a re-id embedding model for the detection and the association stages. The total framework is shown in Figure~\ref{fig:pipeline}.

We design our detection model under anchor-free paradigm~\cite{fcos}. Being benefited from compact model design and simplified parameter settings, the detection branch obtains a satisfactory inference speed. As shown in Figure~\ref{fig:pipeline}, we use DLA-34~\cite{dla} as our model backbone and feature pyramid networks (FPN~\cite{fpn}) as our model neck. We aggregate feature maps by FPN and use multiple-scale features from three levels {P3, P4, P5}, the strides $s$ of features are 8, 16, 32, respectively. The backbone and neck are shared in both template and search branch.

In template branch, we use an efficient feature align method~\cite{roialign} to crop target features. And then a group of controllers that use these features to generate convolution parameters for d-convs. In search branch, following the stacking designs of modern detection methods~\cite{retinanet, fcos}, we embed our d-convs in specific layers to filter the useful features. In re-id embedding branch, we design to generate N-dimensions re-id features for each point in stride-4 feature map. N is set to be 128 in our model. With the discriminative re-id feature, we obtain smooth tracking trajectories.

\subsection{One-shot Detection and Embedding Learning}
We follow the common practice of GIS-based tracker to train a class-agnostic detector with d-convs. Firstly, by using a full convolutional network, each location $(x_{i}^{P}, y_{j}^{P})(P=3, 4, 5)$ on the feature map of different FPN levels can be mapped back onto the original image as

\begin{equation}
\begin{aligned}
(x_{i^{'}}^{ori},y_{j^{'}}^{ori})=(\left \lfloor \frac{s^{P}}{2} \right \rfloor + x_{i}^{P}s^{P}, \left \lfloor \frac{s^{P}}{2} \right \rfloor + y_{j}^{P}s^{P})
\end{aligned}
\label{eq:stride}
\end{equation}

where i and j indicate the $x,y$-coordinates on the feature map, $ori$ indicates original input image. Then we define a center region box on original image as the sampling box $(c_{i}^{ori}-rs^{P},c_{j}^{ori}-rs^{P},c_{i}^{ori}+rs^{P},c_{j}^{ori}+rs^{P})$, where $(c_{i}^{ori},  c_{j}^{ori})$ denotes the annotation center of the target, $r$ is a scale parameter being 1.5, which is the same as the default setting~\cite{fcos}. According to the former definition, we define the location on FPN's feature that can be mapped onto the center region box on original input image as a positive sample $\sigma(=1)$, otherwise a negative sample. Note that we train the regression head and center-ness head only on positive samples. For re-id branch, we treat it as a classification task. As there are no multiple objects annotations in single object tracking dataset, we introduce MOT datasets~\cite{caltech, mot16, cuhk-sysu, prw} to train our re-id branch alternatively. Finally, our model predicts a 2-D vector $\hat{\sigma}$ for classification, a 4-D vector $\hat{\tau}=(\hat{l},\hat{t},\hat{r},\hat{b})$ for bounding box regression, where $(\hat{l},\hat{t},\hat{r},\hat{b})$ indicates the distances from the box center to four sides, a center-ness score $\hat{\varphi}$ for classification regularization and a 128-D re-id embedding feature $\hat{\psi}$. 

\begin{figure}
\begin{center}
\setlength{\fboxrule}{0pt}
\setlength{\fboxsep}{0cm}
\fbox{\includegraphics[width=1\linewidth]{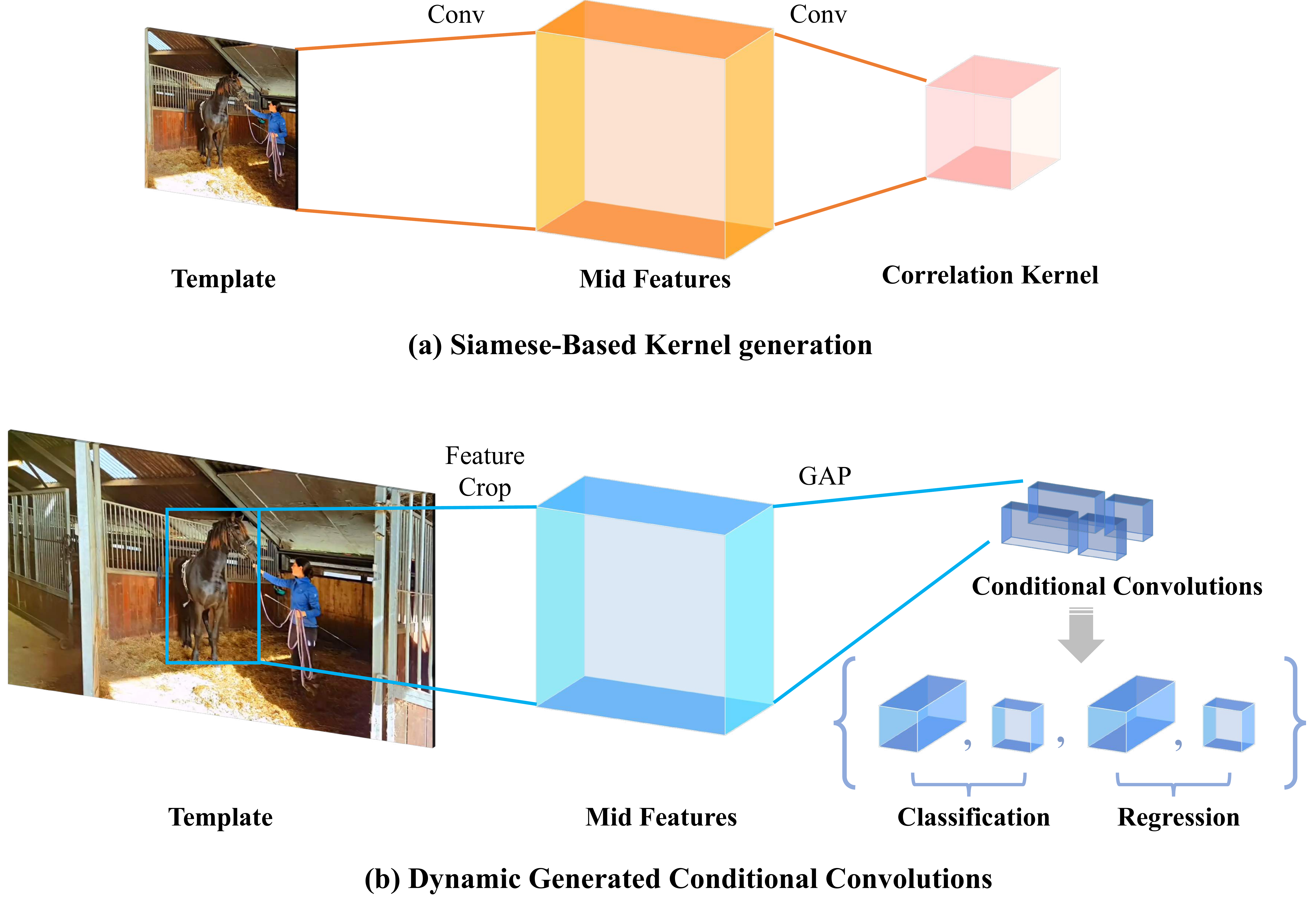}}
\end{center}
  \caption{Comparison of different kernel generation methods. (a) Siamese-based method generates a large kernel by extracting feature from a coarse template image which includes many noises. (b) With aligned feature cropping~\cite{roialign}, convolutions were generated by a global average pooling layer, which are more flexible and effective.}
\label{fig:structure_template}
\end{figure}

\noindent
{\bf Controller Heads.} How to extract abundant target features is an important problem for visual tracking. Nevertheless, this problem is seldom discussed. In Siamese-based methods~\cite{siamfc, siamban, siamrpn++} (as shown in Figure~\ref{fig:structure_template}), the template branch uses a coarse cropping to extract features, which is not appropriate for similar matching due to the noises it involved. Furthermore, the size and the type of the Siamese kernel are fixed for the final activation map which is difficult for model reconstruction. In the dynamic generating method, firstly, we extract the target information from head layers using a feature cropping technology\cite{roialign}. Then with a $1\times1$ $conv$ encoder to adjust the feature channels $\mathcal{C}^{g}$ (as shown in Equation~\ref{eq:params1}) to adapt to the required parameter numbers as shown in Equation~\ref{eq:params2}. Finally, we use a global average pooling layer to generate $\mathcal{C}^{g}-D$ vectors for filter parameters of classification and regression heads.

\vspace{-0.5cm}

\begin{equation}
\begin{aligned}
&\mathcal{C}^{g}=\sum_{u=1}^{p}\mathit{\mathcal{PN}(conv_{cls}^{u})}+\sum_{v=1}^{q}\mathit{\mathcal{PN}(conv_{reg}^{v})}
\end{aligned}
\label{eq:params1}
\end{equation}
\begin{equation}
\begin{aligned}
&\mathcal{PN}(conv_{cls}^{u})=(C^{u}\times K_{\mathit{w}}^{u}\times K_{\mathit{h}}^{u}+1)\times C^{u+1} \\
&\mathcal{PN}(conv_{reg}^{v})=(C^{v}\times K_{\mathit{w}}^{v}\times K_{\mathit{h}}^{v}+1)\times C^{v+1}
\end{aligned}
\label{eq:params2}
\end{equation}
where $p, q$ denote layer numbers counted after model's neck (from 1 to 6). $\mathcal{PN}$ means \textit{parameter numbers} of the d-convs, e.g.\ , in $u$ layer of classification branch, the amount of parameters consist of weights and bias. Specifically, the weight's parameters can be a multiplication of input feature map's channel $C^{u}$, kernel width $K_{w}^{u}$, kernel height $K_{h}^{u}$ and kernel number $C^{u+1}$.

\noindent
{\bf Detection Heads with Dynamic Convolutions.} The detection head contains four components: classification, regression, center-ness and dynamic convolutions. In our model structure, there are four convolutions after neck, an encoder to reduce the channels for efficient computation and prediction layers for each head. We first define numbers of feature layers for classification and regression heads which from 1 to $p$ and $q$, respectively. Then we define groups of layers $\left \{ u_{i}\mid 0\leqslant i\leq p, i\in \mathbb{N^{*}} \right \}$ and $\left \{ v_{j}\mid 0\leqslant j\leq q, j\in \mathbb{N^{*}} \right \}$ to insert our d-convs. By integrating target information to detection head with d-convs, we predict the target and directly regress bounding box at each location on feature maps. Following FCOS~\cite{fcos}, we also predict the center-ness scores associate with regression branch for further robustness.

\noindent
{\bf Re-id Embedding.} We jointly learn detection task and re-id embedding in order to distinguish similar objects. Here we use a convolutional layer on the low level of backbone features to extract re-id embedding features with stride 4. Each re-id feature $E(x, y)\in\mathbb{R}^{C^{E}}$ in location $(x,y)$ present the object whose center annotation that closest to it , $C^{E}$ is the dimension of the embedding feature and set to 128 in our settings.

\begin{table*}[]
\begin{center}
\begin{tabular}{cccc|c|cccc|cccc}
\hline \hline
${\bf CC}$  & ${\bf DW}$ & ${\bf HP}$ & ${\bf DC}$ & ${\bf Layers}$ & ${\bf Top-1}$ & ${\bf PC}$ & ${\bf AC}$ & ${\bf Re-id}$   & ${\bf Success}\uparrow$      & ${\bf Precision}\uparrow$  & ${\bf FPS}\uparrow$ & ${\bf GPU Days}\downarrow$ \\ \hline
 \checkmark & & &            & 1+6             &\checkmark &- &- &-      &49.8 &50.0 &27 &15 \\
            & \checkmark & & & 1+6             &\checkmark &- &- &-      &51.4 &52.4 &32 &9.5 \\
            & &\checkmark &  & 1+6             &\checkmark &- &- &-      &51.1 &52.5 &32 &9.5 \\
            & & & \checkmark & 1+6             &\checkmark &- &- &-      &53.0 &54.2 &32 &10 \\ \hline
            -&- &- &\checkmark & 6               &\checkmark &- &- &-      &48.7 &49.2 &34 &9 \\
            -&- &- &\checkmark & $\left[1\rightarrow6\right]$ &\checkmark &- &- &-      &53.2 &54.3 &28 &19 \\ \hline
            -&- &- & \checkmark & 1+6             & &\checkmark & &      &46.9 &42.4 &32 &10 \\
            -&- &- & \checkmark & 1+6             & & &\checkmark &      &55.2 &55.9 &16 &10 \\
            -&- &- & \checkmark & 1+6             & & & &\checkmark      &57.4 &58.0 &31 &11 \\
 \hline \hline
\end{tabular}
\end{center}
\caption{Ablation studies of different model designs and object association strategies that influences the model's capacity. We also evaluate the time cost for training and inference on a Titan Xp.}
\label{table:ablation}
\end{table*}

\subsection{Loss Function} 
We model the learning task of our framework as a multi-task problem. There are two learning objectives in our full pipeline: detection and re-id. For the detection part, we have three loss functions for classification $\mathcal{L_{\alpha}}$, regression $\mathcal{L_{\beta}}$ and center-ness $\mathcal{L_{\gamma}}$ as following
\begin{equation}
\begin{aligned}
\mathcal{L}_{\mathit{\alpha }}(\sigma _{i,j})=\sum_{i,j}\mathcal{L}_{focal}( \hat{\sigma}_{i,j}, \sigma_{i,j})\label{eq:losscls}
\end{aligned}
\end{equation}
\begin{equation}
\begin{aligned}
\mathcal{L}_{\mathit{\beta }}(\tau_{i,j})=\left\{\begin{matrix}
  \sum_{i,j}\mathcal{L}_{IoU}( \hat{\tau}_{i,j}, \tau_{i,j}) & \sigma_{i,j} = 1 \\ 
0 & otherwise
\end{matrix}\right.
\end{aligned}\label{eq:lossreg}
\end{equation}
\begin{equation}
\begin{aligned}
\mathcal{L}_{\mathit{\gamma }}(\varphi_{i,j})=\left\{\begin{matrix}
  \sum_{i,j}\mathcal{L}_{BCE}( \hat{\varphi}_{i,j}, \varphi_{i,j}) & \sigma_{i,j} = 1 \\ 
0 & otherwise
\end{matrix}\right.
\end{aligned}\label{eq:losscenterness}
\end{equation}

\noindent
The detection loss of multiple scales and heads can be summarized as

\begin{equation}
\begin{aligned}
\mathcal{L}_{\mathit{\det }}=\sum_{m=3,4,5}\sum_{n=\alpha, \beta, \gamma}\lambda_{n}^{m}\mathcal{L}_{n}^{m}\label{eq:lossdet}
\end{aligned}
\end{equation}
where $\lambda_{n}^{m}$ are loss weights to balance these loss functions. Further, we follow the cross-query loss in ~\cite{globaltrack} to improve the discriminating ability of our method as following

\begin{equation}
\begin{aligned}
\mathcal{L}^{'}_{\mathit{\det }}=\frac{1}{I}\sum_{i=1}^{I}\mathcal{L}_{\mathit{\det }}\label{eq:crossqueryloss}
\end{aligned}
\end{equation}
where $I$ indicates the template-search pairs for a pair of images, which means we calculate average loss over different targets in one search image. 

For the re-id embedding part, we treat object identity as a classification problem and use loss function like in ~\cite{fairmot} for the model training 

\begin{equation}
\begin{aligned}
\mathcal{L}_{\mathit{reid }}=\sum_{i=1}^{M}\sum_{m=1}^{J}\mathcal{L}_{\mathit{softmax }}\label{eq:reidloss}
\end{aligned}
\end{equation}

where $J$ is the number of classes, $M$ is the number of objects. And we use strategy in ~\cite{jde} to balance the detection and re-id loss.

\subsection{Online Tracking}
\noindent
{\bf Network Inference.} The inference of our model is straightforward. We initialize the dynamic convolutions using the first frame annotation and keep the re-id feature of the target. Then in subsequent frames, we use the generated kernels to convolve the feature maps in multiple layers. Finally, we take the top-k candidates ordered by the classification score and use a variant non-maximum suppression (NMS) strategy to provide a group of interests.

\noindent
{\bf Online Box Linking.} We use three clues for box linking as in ~\cite{fairmot}: appearance information (\ie re-id features), position information (\ie IoU between adjacent frames) and motion information (\ie Kalman Filter). With these abundant clues, we get smooth box linking with simple Hungarian algorithm~\cite{hungarian}.

\section{Experiments}

\subsection{Implementation Details}
\label{section:training_evaluation}
\noindent
{\bf Parameters.} We use light version of FCOS~\cite{fcos} with DLA-34~\cite{dla} backbone as our base model for one-shot detection. The feature channels are 256 for four stack convolutions and 32 for the encoder behind these. In the template branch, we use RoIAlign~\cite{roialign} with output feature size of 7. Then we use a group of $k$ controllers with global average pooling to generate dynamic convolutions (we set k to 4 as shown in Section~\ref{section:ablation}). In our model, we simply generate $1\times1$ convolutions. In the search branch, we embed d-convs behind neck layer and stack convolutions for both classification head and regression head. In re-id embedding, we use a convolution layer on top of the backbone features with 128 channels, the feature map size is a quarter of size of input image.

\noindent
{\bf Training.} We use the same training data as in ~\cite{globaltrack, fairmot} and use the multi-scale data augmentation by sampling shorter size of input image from 256 to 608 with interval 32. Our model is trained with stochastic gradient descent (SGD) with a starting learning rate of $1\times10^{-3}$. We use 360K training iterations and decreased by 10 at iteration 300K and 340K respectively.

\begin{figure}
\begin{center}
\setlength{\fboxsep}{0cm}
\setlength{\fboxrule}{0pt}
\fbox{\includegraphics[width=1\linewidth]{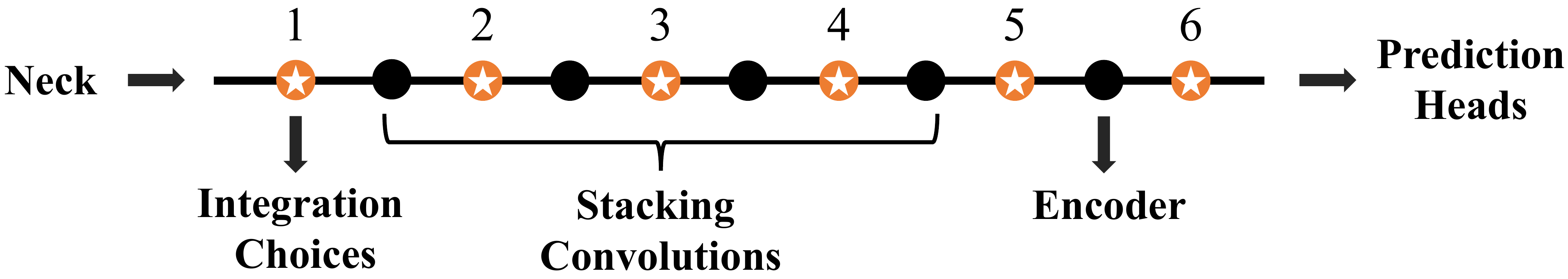}}
\end{center}
  \caption{Numbering the integration layer choices for one-shot detection head.}
\label{fig:ablation}
\end{figure}

\subsection{Ablation Study}
\label{section:ablation}
In this section, we conduct ablation analysis to evaluate different components of our tracker using the LaSOT~\cite{lasot} benchmark. The image size for inference is set to $735\times512$ for all testing.

\noindent
{\bf Effectiveness of Correlation Method.} We make quantitative analysis to compare our dynamic convolutions (DC) generation method with other correlation methods. We denote the cross correlation~\cite{siamfc} by CC, depthwise cross correlation~\cite{siamrpn++} by DW, Hadamard production~\cite{globaltrack} by HP. As shown in Table~\ref{table:ablation}, with similar inference time, we show that d-convs based correlation method are more powerful to model the template information and embed it into search field. This fine-grained feature learning results in strong template correlation. Other methods (i.e. Siamese-based and modulation-based) are special cases of D-Conv. With compact convolutions, we provide a non-trivial solution to unify previous methods, and urge further research on this problem.

\noindent
{\bf Integration Layer Choices.} We experimentally evaluate the influences of integration layer choice, a key factor for correlation capacity. First, we show the sketch of head structures in Figure~\ref{fig:ablation}, there are four stacking convolutions and one encoder, therefore the permutation and combination of six candidate integration layers can result in hundreds choices. Here, we show the relation between the performance and training cost in Table~\ref{table:ablation}. From the table, we can see that with only the high-layer integration (line 5), tracker gets degenerate results. However, with dense connections in stacking layers (line 6), the performance does not boost significantly, but the training cost can be unbearable. In our final model, we integrate d-convs with the \textit{1+6} layers for more practical.

\begin{figure}
\begin{center}
\setlength{\fboxrule}{0pt}
\setlength{\fboxsep}{0cm}
\fbox{\includegraphics[width=1\linewidth]{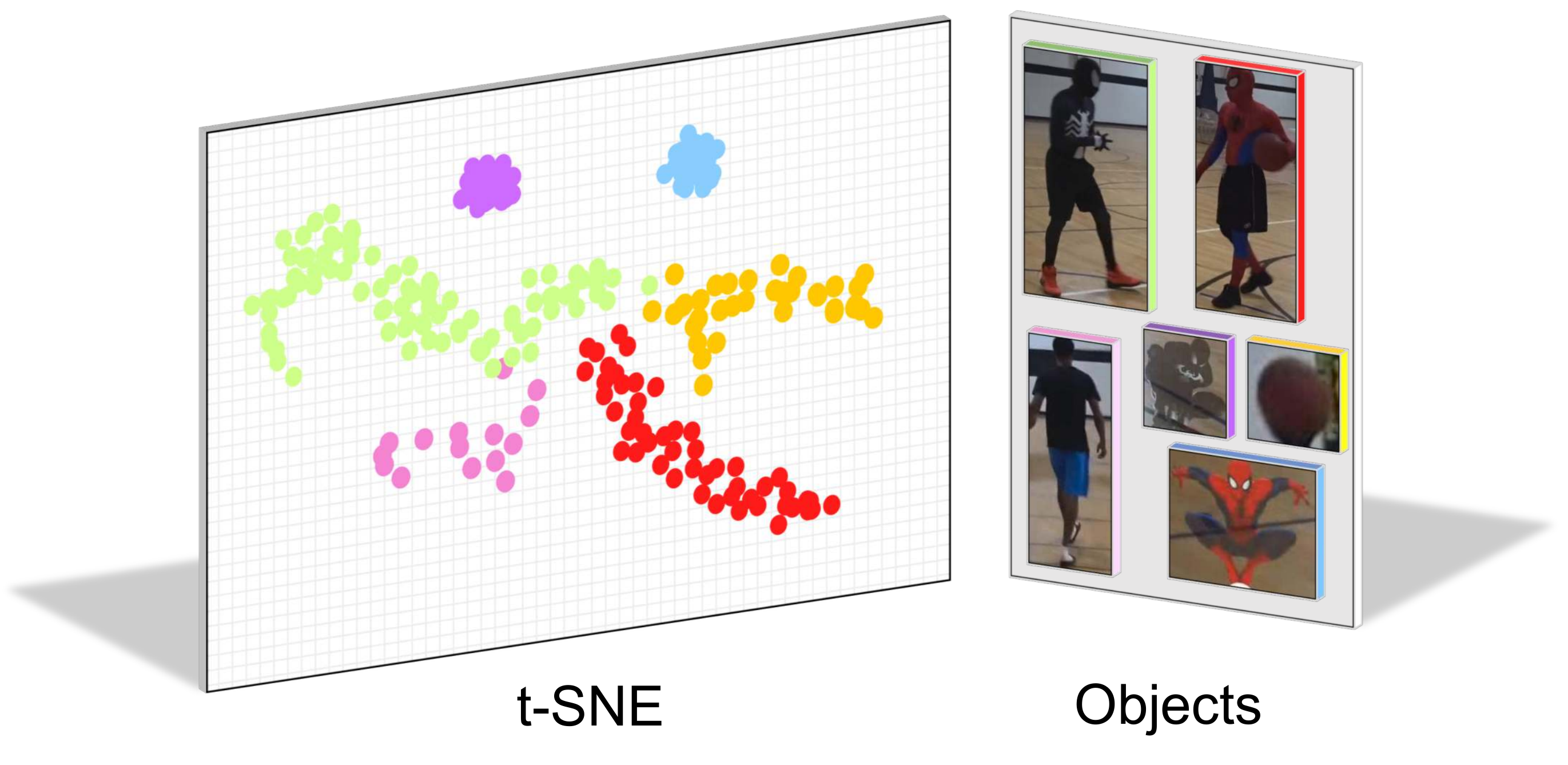}}
\end{center}
  \caption{We show the effectiveness of re-id embedding by using t-SNE~\cite{tsne} to visualize the distance between the features of different objects. The features of the same object are shown by the same color.}
\label{fig:tsne_vis}
\end{figure}

\noindent
{\bf Effectiveness of objects association with re-id embedding.} GIS-based trackers can potentially track all interested objects. These trackers treat the object that have the Top-1 classification score as the target. However, distractor problem leads the tracking performance to deterioration because trackers do not use any constraint. We implement two heuristic constraints to compare with our association method. The first one uses position constraint (denoted by PC). In this constraint, we simply choose the object which is closest to last frame prediction among top 5 candidates. The second one uses appearance constraint (denoted by AC). In this constraint, we use an extra classifier~\cite{dimp} with online update to choose the final target. As shown in Figure~\ref{table:ablation}, our association strategy (denoted by Re-id) outperforms two heuristic methods by a large margin both in precision and computational costs. With only the position or appearance constraint, tracker unable to deal with high-frequency disappearance. However, with an explicit multiple-object association, our tracker is more robust to these real-world challenges. Besides, we use t-SNE~\cite{tsne} technology to show re-id features for different objects. As shown in Figure~\ref{fig:tsne_vis}, we use \textit{person-6} sequence from LaSOT~\cite{lasot} and label six instances in these frames (include \textit{person}, \textit{basketball} and \textit{watermark} on the videos). We show that the re-id embedding can differentiate the inter-class objects and the intra-class objects.

\subsection{Comparison with the state-of-the-art}
\label{section:sota}
\begin{figure}[]
\begin{center}
\setlength{\fboxrule}{0pt}
\setlength{\fboxsep}{0cm}
\fbox{\includegraphics[width=0.47\linewidth]{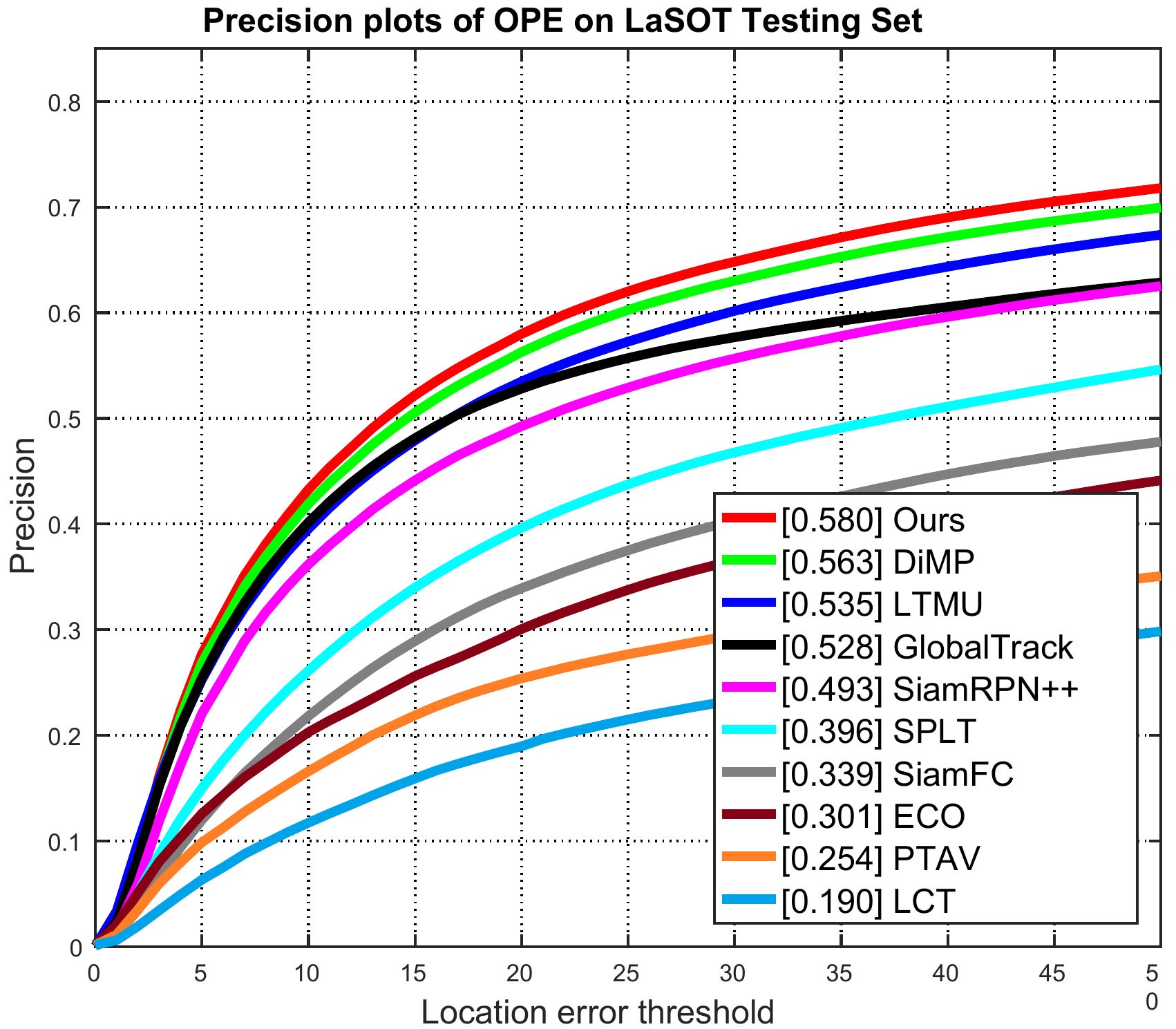}}
\fbox{\includegraphics[width=0.47\linewidth]{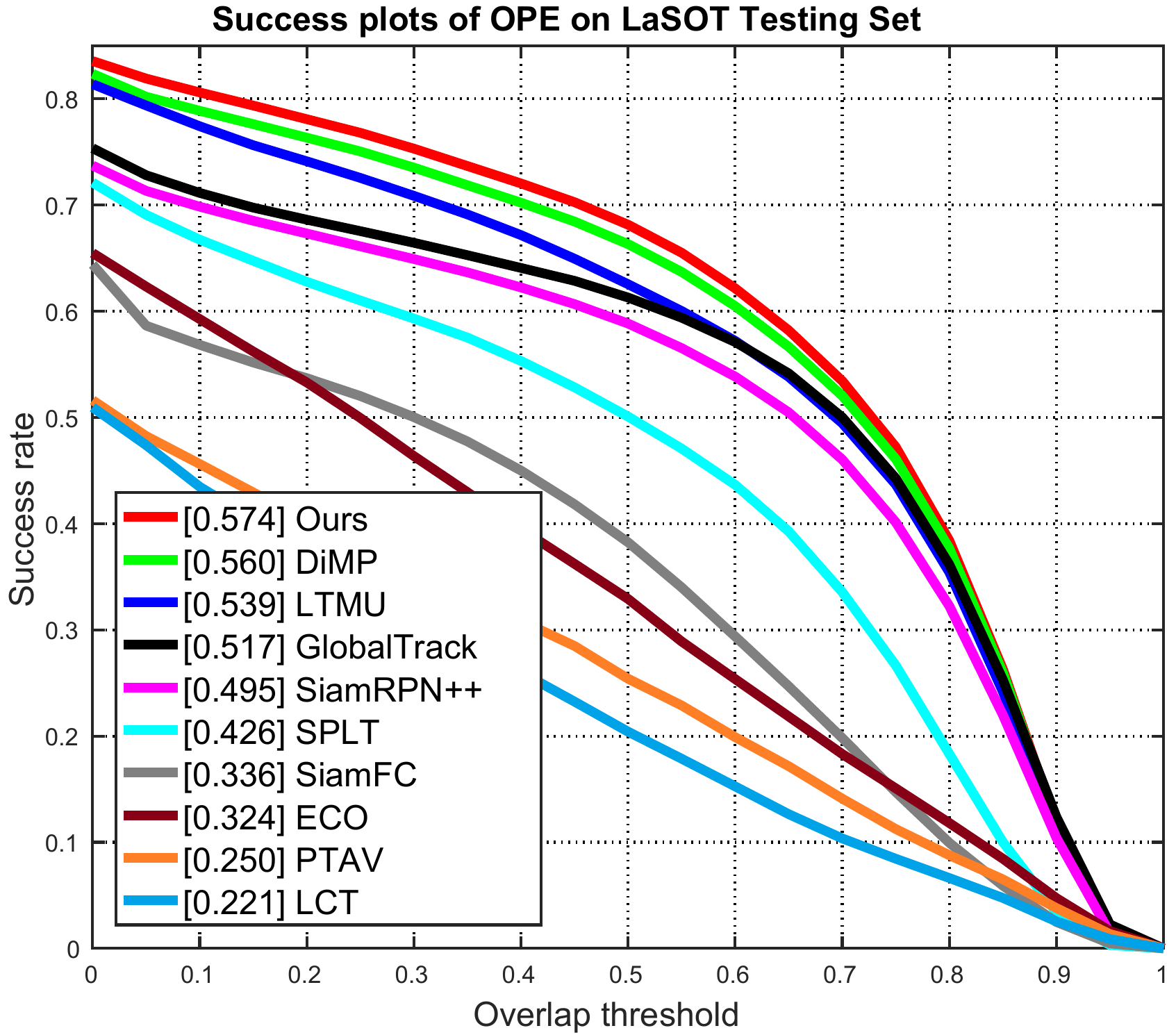}}
\end{center}
  \caption{Plots of ours and state-of-the-art trackers on the test set of LaSOT~\cite{lasot}. Better viewed in color with zoom-in.}
\label{fig:lasot}
\end{figure}

\noindent
{\bf LaSOT.} The LaSOT benchmark~\cite{lasot} is a large-scale modern tracking dataset that contains 1400 long videos (with an average of 2500 frames). In this work, we follow the protocol \uppercase\expandafter{\romannumeral2} defined by official evaluation toolkit and conduct one-pass evaluation with success and precision scores to evaluate our tracker. Compared to nine SOTA methods~\cite{siamfc, dimp, ltmu, eco, ptav, globaltrack, siamrpn++, lct, splt}\footnote {We use the raw result provided by official evaluation toolkit in their website.}, our approach achieves the best results among all competing methods. As shown in Figure~\ref{fig:lasot}, our tracker achieves the best results among all competing methods. Besides that, we maintain a fast inference speed with compact model design, which shows the practicability of our approach. 

\begin{figure}[]
\begin{center}
\setlength{\fboxrule}{0pt}
\setlength{\fboxsep}{0cm}
\fbox{\includegraphics[width=0.47\linewidth]{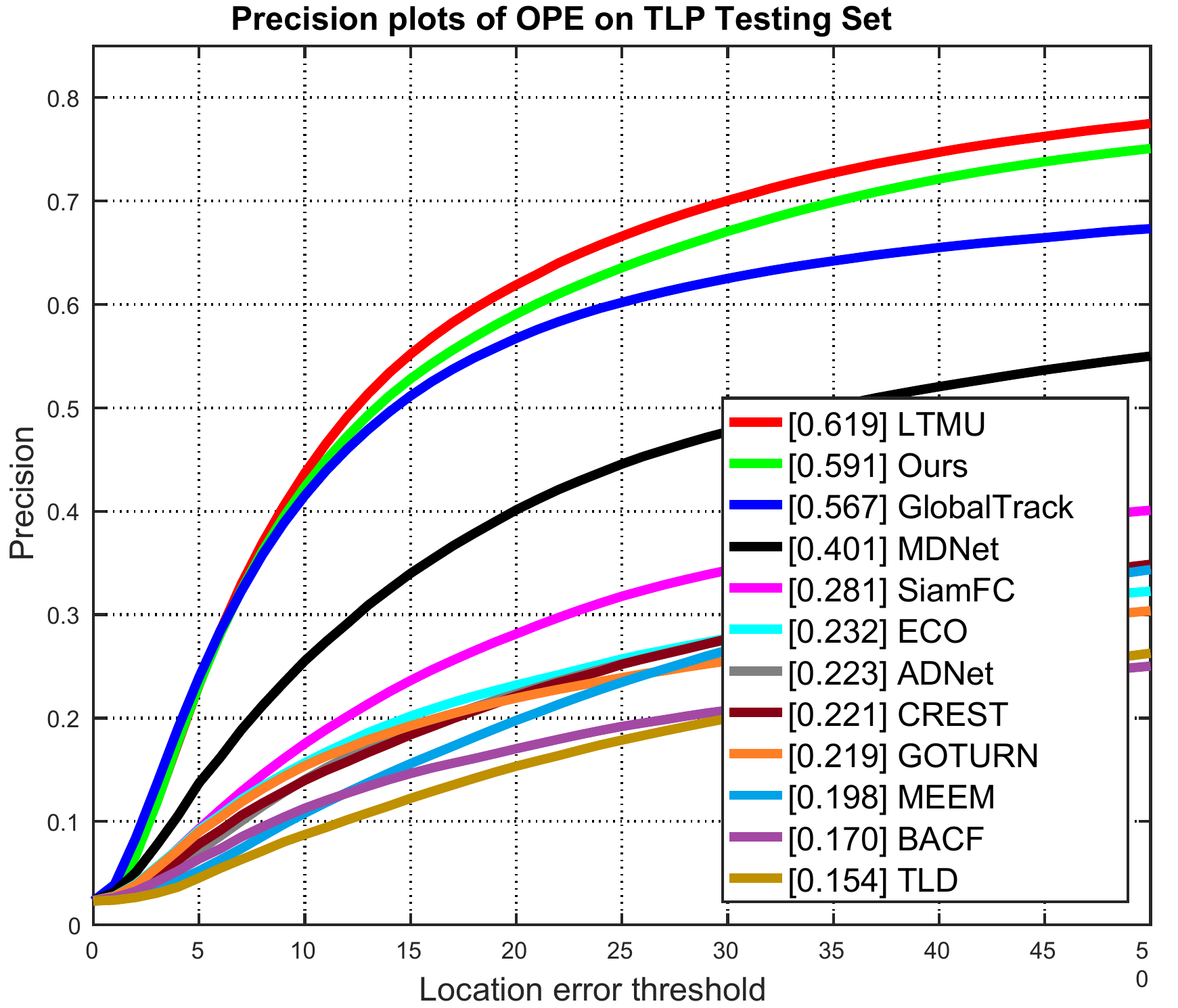}}
\fbox{\includegraphics[width=0.47\linewidth]{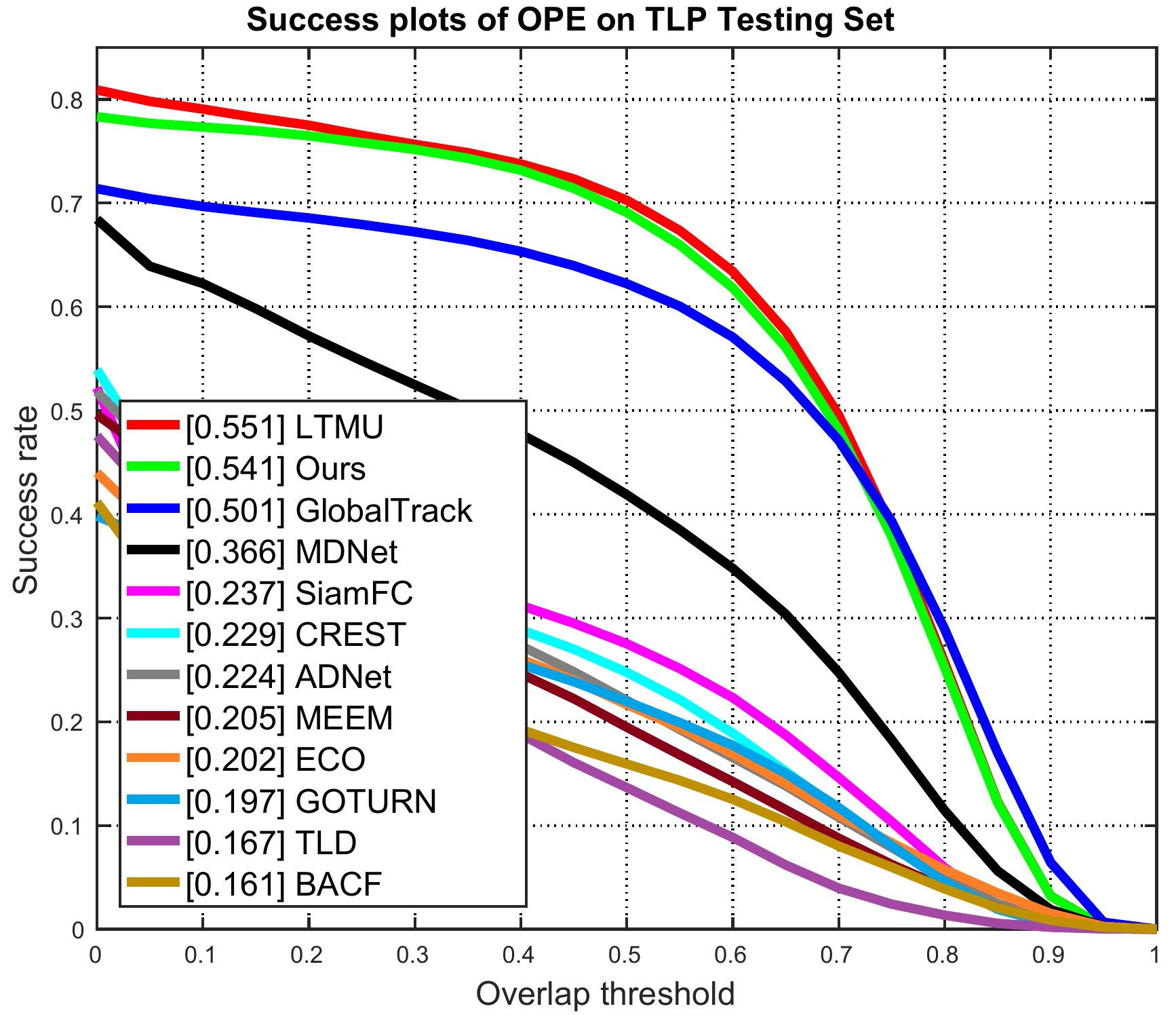}}
\end{center}
  \caption{Plots of Ours and state-of-the-art trackers on the test set of TLP~\cite{tlp}. Better viewed in color with zoom-in.}
\label{fig:tlp}
\end{figure}

\noindent
{\bf OxUvA.} The OxUvA~\cite{oxuva} is a long-term tracking dataset in the wild. The dataset consists of 366 object tracks which are chosen from YTBB~\cite{youtubebb} and labeled at 1Hz frequency. According to the ~\cite{oxuva}, OxUvA is divided into two subset: \textit{dev} and \textit{test}. The test subset contains 166 tracks and each of these lasts for average 2.4 minutes. The evaluation criteria is quite different from short-term benchmarks~\cite{otb2015, vot2019result}, we introduce them briefly as following. The true positive rate ({\bf TPR} calculate the fraction of present objects that are predicted present and precisely. The true negative rate ({\bf TNR}) gives the fraction of absent objects that are determined to disappear. The {\bf MaxGM} provides more convinced measurement to show the trackers performances and is defined as

\vspace{-0.5cm}

\begin{equation}
\begin{split}
\label{eq:oxuva}
\begin{aligned}
&{\bf MaxGM}=\\
&\mathop{max}\limits_{0\leq p\leq 1}\sqrt{((1-p)\cdot {\bf TPR})((1-p)\cdot {\bf TNR}+p)}
\end{aligned}
\end{split}
\end{equation}

We compare our method with eight competing approaches using the open challenge illustrated in ~\cite{oxuva}. In this challenge, trackers can use any public dataset as the training data expect for the YTBB~\cite{youtubebb} validation set. As we can see in Table~\ref{table:oxuva}, our method achieves comparable performance to sophisticated designed long-term trackers that have heavy computation. However, our method runs in real-time, which is practical for applications. 

\begin{table}[]
\begin{center}
\begin{tabular}{cccc|c}
\hline \hline
{\bf (\%)}  & {\bf MaxGM} & {\bf TPR} & {\bf TNR} & {\bf FPS} \\ \hline
 ECO-HC~\cite{eco}      &31.4  &39.5 &0.0  &-   \\
 MDNet~\cite{mdnet}       &34.3  &47.2 &0.0  &1   \\
 LCT~\cite{lct}         &39.6  &29.2 &53.7 &20  \\
 TLD~\cite{tld}         &43.1  &20.8 &\textcolor{red}{89.5} &\textcolor{green}{23}  \\ 
 MBMD~\cite{mbmd}        &54.5  &\textcolor{green}{60.9} &48.5 &3   \\
 GlobalTrack\cite{globaltrack} &60.3  &57.4 &63.3 &10  \\
 SPLT~\cite{splt}        &\textcolor{green}{62.2}  &49.9 &\textcolor{blue}{77.6} &\textcolor{blue}{27}  \\
 Siam R-CNN~\cite{siamr-cnn}        &\textcolor{red}{72.3}  &\textcolor{red}{70.1} &\textcolor{green}{74.5} &4.7  \\ \hline
 {\bf Ours}  &\textcolor{blue}{68.8}      &\textcolor{blue}{68.6}     &69.4     &\textcolor{red}{31}\\
 \hline \hline
\end{tabular}
\end{center}
\caption{State-of-the-art comparison on the test set of OxUvA~\cite{oxuva} in terms of MaxGM, TPR and TNR. The best three results are shown in \textcolor{red}{red}, \textcolor{blue}{blue} and \textcolor{green}{green} colors, respectively.}
\label{table:oxuva}
\end{table}

\noindent
{\bf TLP.} The TLP is a long video dataset for object tracking. The dataset including 50 long videos of 676K frames (over 400 minutes). We follow the OPE evaluation that used in ~\cite{otb2015} and compare our tracker with other trackers. As shown in Figure~\ref{fig:tlp}, our tracker outperforms another GIS-based tracker\cite{globaltrack} by a large margin and gets a comparable performance to the best tracker\cite{ltmu} in this benchmark. Compared with another GIS-based tracker GlobalTrack~\cite{globaltrack}, our model get a good balance between precision and recall.

\noindent
{\bf VOT2018LT.} We compare our tracker with other state-of-the art tracking algorithms on VOT2018LT benchmark~\cite{vot2018result}. In this dataset, there are 35 long videos with 146K frames in total. The challengs in these sequences are varied, including long-term target disappearances and severe occlusion, which require trackers to be more robust. The evaluation criterion of VOT2018LT dataset includes tracking precision ({\bf Pr}), tracking recall ({\bf Re}) and tracking {\bf F-score}. We report the tracking performance of our tracker and other competing ones in Table~\ref{table:votlt}. As we see, our tracker achieves an absolute gain of 5\% in terms of F-score. The results demonstrate the strong performance of our approach in long-term tracking scenarios.

\begin{table}[]
\fontsize{5.5}{5.5}\selectfont
\renewcommand\arraystretch{1.35}
\begin{tabular}{cccc|cccc}
\hline \hline
\multicolumn{4}{c|}{\textbf{VOT2018LT}}                         & \multicolumn{4}{c}{\textbf{VOT2019LT}}                          \\ \hline 
\textbf{Tracker} & \textbf{F-score} & \textbf{Pr} & \textbf{Re} & \textbf{Tracker} & \textbf{F-score} & \textbf{Pr} & \textbf{Re}  \\ \hline
PTAVplus         & 0.481            & 0.595       & 0.404   & FuCoLoT    & 0.411            & 0.507       & 0.346                 \\
SYT              & 0.509            & 0.520       & 0.499   & ASINT    & 0.505            & 0.517       & 0.494                   \\
LTSINT           & 0.536            & 0.566       & 0.510   & CooSiam    & 0.508            & 0.482       & 0.537                 \\
MMLT             & 0.546            & 0.574       & 0.521   & SiamRPNsLT    & 0.556            &\textcolor{red}{0.749}       & 0.443              \\
DaSiam\_LT       & 0.607            & 0.627       & 0.588   & mbdet    & 0.567            & 0.609       & 0.530                   \\
MBMD             & 0.610            &\textcolor{green}{0.634}       & 0.588    & SiamDW\_LT    & 0.665            & 0.697       &\textcolor{green}{0.636}            \\
SPLT             &\textcolor{green}{0.616}            &0.633       &\textcolor{green}{0.600}   & CLGS     &\textcolor{green}{0.674}            &\textcolor{blue}{0.739}       & 0.619                   \\
SiamRPN++        &\textcolor{blue}{0.629}            &\textcolor{blue}{0.649}       &\textcolor{blue}{0.609}   & LT\_DSE     &\textcolor{red}{0.695}            &\textcolor{green}{0.715}       &\textcolor{red}{0.677}                \\ \hline
\textbf{Ours}    &\textcolor{red}{0.683}            &\textcolor{red}{0.687}       &\textcolor{red}{0.655}    &\textbf{Ours}    &\textcolor{blue}{0.687}    & 0.690      &\textcolor{blue}{0.662}                 \\ \hline \hline
\end{tabular}
\vspace{6.5pt}
\caption{State-of-the-art comparison on the VOT2018LT~\cite{vot2018result} and VOT2019LT~\cite{vot2019result} benchmarks in terms of F-score, Pr and Re. The best three results are shown in \textcolor{red}{red}, \textcolor{blue}{blue} and \textcolor{green}{green} colors.}
\label{table:votlt}
\end{table}

\noindent
{\bf VOT2019LT.} The VOT2019LT benchmark~\cite{vot2019result} is an extension of the 2018 version~\cite{vot2018result} that contains 50 challenging sequences. Each video contains 10 long-range disappearances on average. The evaluation protocol is similar to that in VOT2018LT~\cite{vot2018result}. Table~\ref{table:votlt} show that our model get a promising result compared to the well-designed long-term trackers for the competition. The tracking results demonstrate the advantage of GIS-based paradigm.

\section{Conclusions}
In this work, we propose a new long-term tracking paradigm which consists of one-shot detection and object association. To achieve an efficient detection model, we design a novel dynamic convolutions generation method for flexible feature correlation. Further, in order to distinguish the target from distractors, we present a compact object association strategy with discriminative re-id embedding. Numerous experiments on five long-term tracking benchmarks verify the performance of the proposed approach. Potentiated by its efficiency, we believe that the proposed framework can be performed as a new baseline for further studies.

\noindent
\textbf{Acknowledgments.}

This work was supported by the National Natural Science Foundation of China (No. 61972167, 61802135, 61872112), the Project of Guangxi Science and Technology (No. 2020AC19194), the Guangxi “Bagui Scholar” Teams for Innovation and Research Project, the Guangxi Collaborative Innovation Center of Multi-source Information Integration and Intelligent Processing, the Guangxi Talent Highland Project of Big Data Intelligence and Application, and the Open Project Program of the National Laboratory of Pattern Recognition (NLPR) (No. 202000012).

{\small
\bibliographystyle{ieee_fullname}
\bibliography{Submission-2349}

\begin{thebibliography}{10}\itemsep=-1pt

\bibitem{siamfc}
Luca Bertinetto, Jack Valmadre, Jo{\~{a}}o~F. Henriques, Andrea Vedaldi, and
  Philip H.~S. Torr.
\newblock Fully-convolutional siamese networks for object tracking.
\newblock In {\em {ECCV} Workshops {(2)}}, volume 9914 of {\em Lecture Notes in
  Computer Science}, pages 850--865, 2016.

\bibitem{dimp}
Goutam Bhat, Martin Danelljan, Luc~Van Gool, and Radu Timofte.
\newblock Learning discriminative model prediction for tracking.
\newblock In {\em {ICCV}}, pages 6181--6190. {IEEE}, 2019.

\bibitem{siamban}
Zedu Chen, Bineng Zhong, Guorong Li, Shengping Zhang, and Rongrong Ji.
\newblock Siamese box adaptive network for visual tracking.
\newblock In {\em {CVPR}}, pages 6667--6676. {IEEE}, 2020.

\bibitem{ltmu}
Kenan Dai, Yunhua Zhang, Dong Wang, Jianhua Li, Huchuan Lu, and Xiaoyun Yang.
\newblock High-performance long-term tracking with meta-updater.
\newblock In {\em {CVPR}}, pages 6297--6306. {IEEE}, 2020.

\bibitem{eco}
Martin Danelljan, Goutam Bhat, Fahad~Shahbaz Khan, and Michael Felsberg.
\newblock {ECO:} efficient convolution operators for tracking.
\newblock In {\em {CVPR}}, pages 6931--6939. {IEEE} Computer Society, 2017.

\bibitem{srdcf}
Martin Danelljan, Gustav H{\"{a}}ger, Fahad~Shahbaz Khan, and Michael Felsberg.
\newblock Learning spatially regularized correlation filters for visual
  tracking.
\newblock In {\em {ICCV}}, pages 4310--4318. {IEEE} Computer Society, 2015.

\bibitem{caltech}
Piotr Doll{\'{a}}r, Christian Wojek, Bernt Schiele, and Pietro Perona.
\newblock Pedestrian detection: {A} benchmark.
\newblock In {\em {CVPR}}, pages 304--311. {IEEE} Computer Society, 2009.

\bibitem{lasot}
Heng Fan, Liting Lin, Fan Yang, Peng Chu, Ge Deng, Sijia Yu, Hexin Bai, Yong
  Xu, Chunyuan Liao, and Haibin Ling.
\newblock Lasot: {A} high-quality benchmark for large-scale single object
  tracking.
\newblock In {\em {CVPR}}, pages 5374--5383. Computer Vision Foundation /
  {IEEE}, 2019.

\bibitem{ptav}
Heng Fan and Haibin Ling.
\newblock Parallel tracking and verifying.
\newblock {\em {IEEE} Trans. Image Process.}, 28(8):4130--4144, 2019.

\bibitem{roialign}
Kaiming He, Georgia Gkioxari, Piotr Doll{\'{a}}r, and Ross~B. Girshick.
\newblock Mask {R-CNN}.
\newblock In {\em {ICCV}}, pages 2980--2988. {IEEE} Computer Society, 2017.

\bibitem{gapdt}
Lianghua Huang, Xin Zhao, and Kaiqi Huang.
\newblock Bridging the gap between detection and tracking: {A} unified
  approach.
\newblock In {\em {ICCV}}, pages 3998--4008. {IEEE}, 2019.

\bibitem{globaltrack}
Lianghua Huang, Xin Zhao, and Kaiqi Huang.
\newblock Globaltrack: {A} simple and strong baseline for long-term tracking.
\newblock In {\em {AAAI}}, pages 11037--11044. {AAAI} Press, 2020.

\bibitem{dynamicfilter}
Xu Jia, Bert~De Brabandere, Tinne Tuytelaars, and Luc~Van Gool.
\newblock Dynamic filter networks.
\newblock In {\em {NIPS}}, pages 667--675, 2016.

\bibitem{rtmdnet}
Ilchae Jung, Jeany Son, Mooyeol Baek, and Bohyung Han.
\newblock Real-time mdnet.
\newblock In {\em {ECCV} {(4)}}, volume 11208 of {\em Lecture Notes in Computer
  Science}, pages 89--104. Springer, 2018.

\bibitem{tld}
Zdenek Kalal, Krystian Mikolajczyk, and Jiri Matas.
\newblock Tracking-learning-detection.
\newblock {\em {IEEE} Trans. Pattern Anal. Mach. Intell.}, 34(7):1409--1422,
  2012.

\bibitem{hungarian}
Harold~W. Kuhn.
\newblock The hungarian method for the assignment problem.
\newblock In {\em 50 Years of Integer Programming}, pages 29--47. Springer,
  2010.

\bibitem{siamrpn++}
Bo Li, Wei Wu, Qiang Wang, Fangyi Zhang, Junliang Xing, and Junjie Yan.
\newblock Siamrpn++: Evolution of siamese visual tracking with very deep
  networks.
\newblock In {\em {CVPR}}, pages 4282--4291. Computer Vision Foundation /
  {IEEE}, 2019.

\bibitem{fpn}
Tsung{-}Yi Lin, Piotr Doll{\'{a}}r, Ross~B. Girshick, Kaiming He, Bharath
  Hariharan, and Serge~J. Belongie.
\newblock Feature pyramid networks for object detection.
\newblock In {\em {CVPR}}, pages 936--944. {IEEE} Computer Society, 2017.

\bibitem{retinanet}
Tsung{-}Yi Lin, Priya Goyal, Ross~B. Girshick, Kaiming He, and Piotr
  Doll{\'{a}}r.
\newblock Focal loss for dense object detection.
\newblock In {\em {ICCV}}, pages 2999--3007. {IEEE} Computer Society, 2017.

\bibitem{vot2020-1}
Alan Luke{\'z}i{\v{c}}, Luka~{\v{C}}ehovin Zajc, Tom{\'a}{\v{s}}
  Voj{\'\i}{\v{r}}, Ji{\v{r}}{\'\i} Matas, and Matej Kristan.
\newblock Performance evaluation methodology for long-term single-object
  tracking.
\newblock {\em IEEE Transactions on Cybernetics}, 2020.

\bibitem{vot2018result}
Kristan M, Matas J, and et~al. Leonardis~A.
\newblock The sixth visual object tracking vot2018 challenge results.
\newblock In {\em {ECCV} Workshops {(1)}}, volume 11129 of {\em Lecture Notes
  in Computer Science}, pages 3--53. Springer, 2018.

\bibitem{vot2019result}
Kristan M, Matas J, and et~al. Leonardis~A.
\newblock The seventh visual object tracking {VOT2019} challenge results.
\newblock In {\em {ICCV} Workshops}, pages 2206--2241. {IEEE}, 2019.

\bibitem{lct}
Chao Ma, Xiaokang Yang, Chongyang Zhang, and Ming{-}Hsuan Yang.
\newblock Long-term correlation tracking.
\newblock In {\em {CVPR}}, pages 5388--5396. {IEEE} Computer Society, 2015.

\bibitem{tsne}
Laurens van~der Maaten and Geoffrey Hinton.
\newblock Visualizing data using t-sne.
\newblock {\em Journal of machine learning research}, 9(Nov):2579--2605, 2008.

\bibitem{mot16}
Anton Milan, Laura Leal{-}Taix{\'{e}}, Ian~D. Reid, Stefan Roth, and Konrad
  Schindler.
\newblock {MOT16:} {A} benchmark for multi-object tracking.
\newblock {\em CoRR}, abs/1603.00831, 2016.

\bibitem{tlp}
Abhinav Moudgil and Vineet Gandhi.
\newblock Long-term visual object tracking benchmark.
\newblock In {\em {ACCV} {(2)}}, volume 11362 of {\em Lecture Notes in Computer
  Science}, pages 629--645. Springer, 2018.

\bibitem{mdnet}
Hyeonseob Nam and Bohyung Han.
\newblock Learning multi-domain convolutional neural networks for visual
  tracking.
\newblock In {\em {CVPR}}, pages 4293--4302. {IEEE} Computer Society, 2016.

\bibitem{youtubebb}
Esteban Real, Jonathon Shlens, Stefano Mazzocchi, Xin Pan, and Vincent
  Vanhoucke.
\newblock Youtube-boundingboxes: {A} large high-precision human-annotated data
  set for object detection in video.
\newblock In {\em {CVPR}}, pages 7464--7473. {IEEE} Computer Society, 2017.

\bibitem{condinst}
Zhi Tian, Chunhua Shen, and Hao Chen.
\newblock Conditional convolutions for instance segmentation.
\newblock {\em CoRR}, abs/2003.05664, 2020.

\bibitem{fcos}
Zhi Tian, Chunhua Shen, Hao Chen, and Tong He.
\newblock {FCOS:} {A} simple and strong anchor-free object detector.
\newblock {\em CoRR}, abs/2006.09214, 2020.

\bibitem{oxuva}
Jack Valmadre, Luca Bertinetto, Jo{\~{a}}o~F. Henriques, Ran Tao, Andrea
  Vedaldi, Arnold W.~M. Smeulders, Philip H.~S. Torr, and Efstratios Gavves.
\newblock Long-term tracking in the wild: {A} benchmark.
\newblock In {\em {ECCV} {(3)}}, volume 11207 of {\em Lecture Notes in Computer
  Science}, pages 692--707. Springer, 2018.

\bibitem{siamr-cnn}
Paul Voigtlaender, Jonathon Luiten, Philip H.~S. Torr, and Bastian Leibe.
\newblock Siam {R-CNN:} visual tracking by re-detection.
\newblock In {\em {CVPR}}, pages 6577--6587. {IEEE}, 2020.

\bibitem{solov2}
Xinlong Wang, Rufeng Zhang, Tao Kong, Lei Li, and Chunhua Shen.
\newblock Solov2: Dynamic, faster and stronger.
\newblock {\em CoRR}, abs/2003.10152, 2020.

\bibitem{jde}
Zhongdao Wang, Liang Zheng, Yixuan Liu, and Shengjin Wang.
\newblock Towards real-time multi-object tracking.
\newblock {\em CoRR}, abs/1909.12605, 2019.

\bibitem{otb2015}
Yi Wu, Jongwoo Lim, and Ming{-}Hsuan Yang.
\newblock Online object tracking: {A} benchmark.
\newblock In {\em {CVPR}}, pages 2411--2418. {IEEE} Computer Society, 2013.

\bibitem{cuhk-sysu}
Tong Xiao, Shuang Li, Bochao Wang, Liang Lin, and Xiaogang Wang.
\newblock Joint detection and identification feature learning for person
  search.
\newblock In {\em {CVPR}}, pages 3376--3385. {IEEE} Computer Society, 2017.

\bibitem{splt}
Bin Yan, Haojie Zhao, Dong Wang, Huchuan Lu, and Xiaoyun Yang.
\newblock 'skimming-perusal' tracking: {A} framework for real-time and robust
  long-term tracking.
\newblock In {\em {ICCV}}, pages 2385--2393. {IEEE}, 2019.

\bibitem{condconv}
Brandon Yang, Gabriel Bender, Quoc~V. Le, and Jiquan Ngiam.
\newblock Condconv: Conditionally parameterized convolutions for efficient
  inference.
\newblock In {\em NeurIPS}, pages 1305--1316, 2019.

\bibitem{dla}
Fisher Yu, Dequan Wang, Evan Shelhamer, and Trevor Darrell.
\newblock Deep layer aggregation.
\newblock In {\em {CVPR}}, pages 2403--2412. {IEEE} Computer Society, 2018.

\bibitem{fairmot}
Yifu Zhang, Chunyu Wang, Xinggang Wang, Wenjun Zeng, and Wenyu Liu.
\newblock A simple baseline for multi-object tracking.
\newblock {\em CoRR}, abs/2004.01888, 2020.

\bibitem{mbmd}
Yunhua Zhang, Dong Wang, Lijun Wang, Jinqing Qi, and Huchuan Lu.
\newblock Learning regression and verification networks for long-term visual
  tracking.
\newblock {\em CoRR}, abs/1809.04320, 2018.

\bibitem{siamdw}
Zhipeng Zhang and Houwen Peng.
\newblock Deeper and wider siamese networks for real-time visual tracking.
\newblock In {\em {CVPR}}, pages 4591--4600. Computer Vision Foundation /
  {IEEE}, 2019.

\bibitem{ocean}
Zhipeng Zhang, Houwen Peng, Jianlong Fu, Bing Li, and Weiming Hu.
\newblock Ocean: Object-aware anchor-free tracking.
\newblock In Andrea Vedaldi, Horst Bischof, Thomas Brox, and Jan{-}Michael
  Frahm, editors, {\em Computer Vision - {ECCV} 2020 - 16th European
  Conference, Glasgow, UK, August 23-28, 2020, Proceedings, Part {XXI}}, volume
  12366 of {\em Lecture Notes in Computer Science}, pages 771--787. Springer,
  2020.

\bibitem{prw}
Liang Zheng, Hengheng Zhang, Shaoyan Sun, Manmohan Chandraker, Yi Yang, and Qi
  Tian.
\newblock Person re-identification in the wild.
\newblock In {\em {CVPR}}, pages 3346--3355. {IEEE} Computer Society, 2017.

\bibitem{bn-tracking}
Bineng Zhong, Bing Bai, Jun Li, Yulun Zhang, and Yun Fu.
\newblock Hierarchical tracking by reinforcement learning-based searching and
  coarse-to-fine verifying.
\newblock {\em {IEEE} Trans. Image Process.}, 28(5):2331--2341, 2019.

\bibitem{bn-reid}
Qinqin Zhou, Bineng Zhong, Xiangyuan Lan, Gan Sun, Yulun Zhang, Baochang Zhang,
  and Rongrong Ji.
\newblock Fine-grained spatial alignment model for person re-identification
  with focal triplet loss.
\newblock {\em {IEEE} Trans. Image Process.}, 29:7578--7589, 2020.

\bibitem{dasiam}
Zheng Zhu, Qiang Wang, Bo Li, Wei Wu, Junjie Yan, and Weiming Hu.
\newblock Distractor-aware siamese networks for visual object tracking.
\newblock In {\em {ECCV} {(9)}}, volume 11213 of {\em Lecture Notes in Computer
  Science}, pages 103--119. Springer, 2018.

\end{thebibliography}
}

\end{document}